\documentclass[14pt]{article}
\usepackage[utf8]{inputenc}
\usepackage[final]{nips_2018}
\usepackage{wrapfig}
\usepackage{graphicx}
\usepackage{titling}
\usepackage{subcaption}
\usepackage{adjustbox}
\setcitestyle{numbers}
\title{ BERTQA - Attention on Steroids}
\author{Ankit Chadha \\ ankitrc@stanford.edu \and Rewa Sood \\ rrsood@stanford.edu}
\date{March 17, 2019}

\begin{document}
\maketitle

\begin{abstract}
In this work, we extend the Bidirectional Encoder Representations from Transformers (BERT) with an emphasis on directed coattention to obtain an improved F1 performance on the SQUAD2.0 dataset. The Transformer architecture on which BERT is based places hierarchical global attention on the concatenation of the context and query. Our additions to the BERT architecture augment this attention with a more focused context to query (C2Q) and query to context (Q2C) attention via a set of modified Transformer encoder units. In addition, we explore adding convolution based feature extraction within the coattention architecture to add localized information to self-attention. We found that coattention significantly improves the no answer F1 by 4 points in the base and 1 point in the large architecture. After adding skip connections the no answer F1 improved further without causing an additional loss in has answer F1. The addition of localized feature extraction added to attention produced an overall dev F1 of 77.03 in the base architecture. We applied our findings to the large BERT model which contains twice as many layers and further used our own augmented version of the SQUAD 2.0 dataset created by back translation, which we have named SQUAD 2.Q. Finally, we performed hyperparameter tuning and ensembled our best models for a final F1/EM of 82.317/79.442 (Attention on Steroids, PCE Test Leaderboard).
\end{abstract}

\section{Introduction}
Through this CS224N Pre-trained Contextual Embeddings (PCE) project, we tackle the question answering problem which is one of the most popular in NLP and has been brought to the forefront by datasets such as SQUAD 2.0. This problem's success stems from both the challenge it presents and the recent successes in approaching human level function. As most, if not all, of the problems humans solve every day can be posed as a question, creating an deep learning based solution that has access to the entire internet is a critical milestone for NLP. Through our project, our group had tested the limits of applying attention in BERT \cite{BERT} to improving the network's performance on the SQUAD2.0 dataset \cite{SQUAD}. BERT applies attention to the concatenation of the query and context vectors and thus attends these vectors in a global fashion. We propose BERTQA \cite{bertqa} which adds Context-to-Query (C2Q) and Query-to-Context (Q2C) attention in addition to localized feature extraction via 1D convolutions. We implemented the additions ourselves, while the Pytorch baseline BERT code was obtained from \cite{github}. The SQUAD2.0 answers span from a length of zero to multiple words and this additional attention provides hierarchical information that will allow the network to better learn to detect answer spans of varying sizes. We applied the empirical findings from this part of our project to the large BERT model, which has twice as many layers as the base BERT model. We also augmented the SQUAD2.0 dataset with additional backtranslated examples. This augmented dataset will be publicly available on our github \cite{Ank_git} on the completion of this course. After performing hyperparameter tuning, we ensembled our two best networks to get F1 and EM scores of 82.317 and 79.442 respectively. The experiments took around 300 GPU hours to train.

\section{Related Work}
The SQUAD2.0 creators proposed this dataset as a means for networks to actually understand the text they were being interrogated about rather than simply being extractive parsers. Many networks stepped up to the challenge including BERT, BIDAF, and QANET. BERT is a fully feed forward network that is based on the transformer architecture \cite{TX}. The base BERT model has 12 transformer encoder layers that terminate in an interchangeable final layer which can be finetuned to the specific task. We chose this network as our baseline because of its use of contextual embeddings and global attention and because of the speed advantage derived from an RNN free architecture. We derived inspiration for our modifications from the BIDAF and QANET models. BIDAF is an LSTM based network that uses character, word, and contextual embeddings which are fed through Context-to-Query (C2Q) and Query-to-Context (Q2C) layers. The final logits are derived from separate Start and End output layers, as opposed to BERT which produces these logits together. Our C2Q/Q2C addition to BERT and the Dense Layer/LSTM based separate final Start and End logit prediction layer were inspired by this paper. We also refered to the QANET model, which is also a fully feed forward network that emphasizes the use of convolutions to capture the local structure of text. Based on this paper, we created a convolutional layer within the C2Q/Q2C architecture to add localized information to BERT's global attention and the C2Q/Q2C coattention.

In addition to referencing these papers that helped us build a successful model, we also explored many other papers which either didn't work with our transformer based model or simply didn't work in combination with our additions to BERT. The three main papers from which we tried to gain ideas are U-Net: Machine Reading Comprehension with Unanswerable Questions \cite{unet}, Attention-over-Attention Neural Networks for Reading Comprehension \cite{AoA}, and FlowQA: Grasping Flow in History for Conversational Machine Comprehension \cite{flow}. We tried implementing the multitask learning methodology presented in U-Net by passing the [CLS] token through a series of convolutional layers to create a probability of whether the question has an answer. We combined this prediction with the prediction of Start and End logits by combining the logits' crossentropy loss and the [CLS] binary crossentropy loss. Unfortunately, this additional loss seemed to be hindering the network's learning ability. We conjecture that this type of multitask learning would benefit from full training instead of the finetuning we were restricted to doing because of resources and time. We looked to Attention-over-Attention as a source of additional ways of injecting attention into our network. Attention-over-Attention has a dot-product based attention mechanism that attends to attention vectors instead of embedding vectors. We believe this method did not help in our case because BERT works with the Context and Query  as part of the same vector while the Attention-over-Attention model requires completely uncoupled Context and Query vectors. As a side note, we do separate the Context and Query vector derived from BERT before the coattention layers of our model, but these layers are not negatively affected by the fact that these separated vectors contain 'mixed' information between the Context and Query. Finally, we explored the FlowQA paper which proposed combining embeddings from multiple layers as an input to the final prediction layer. We implemented this idea by combining embeddings from multiple BERT layers as an input to our final prediction layer. This final layer was simply an additional transformer encoder and we think that the encoder does not have the LSTM's ability of being able to aggregate information from multiple sources. 

\section{Methods} 
We first focused on directed coattention via context to query and query to context attention as discussed in BIDAF \cite{BIDAF}. We then implemented localized feature extraction by 1D convolutions to add local information to coattention based on the QANET architecture \cite{QANET}. Subsequently, we experimented with different types of skip connections to inject BERT embedding information back into our modified network. We then applied what we learned using the base BERT model to the large BERT model. Finally, we performed hyperparameter tuning by adjusting the number of coattention blocks, the batch size, and the number of epochs trained and ensembled our three best networks. Each part of the project is discussed further in the subsections below.

\subsection{ BERTQA - Directed Coattention}
\begin{figure}[h!]
    \centering
    \includegraphics[height=4.75in]{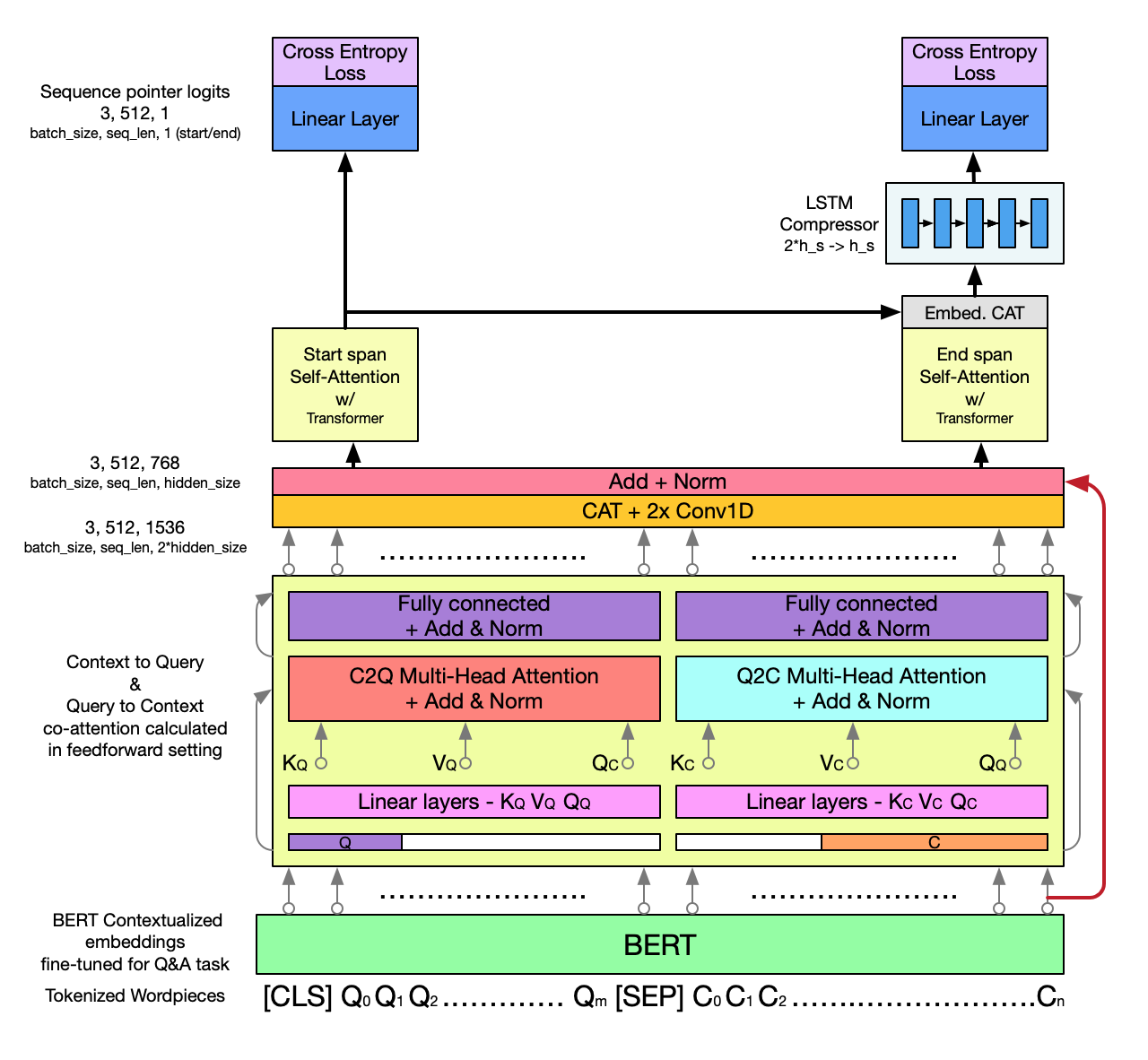}
    \caption{\label{fig:newarch} Proposed C2Q and Q2C directed coattention architecture}
\end{figure}

The base BERT network, the baseline for this project, is built with 12 Transformer encoder blocks. These encoder blocks contain multi-head attention and a feed forward network. Each head of the multi-head attention attends to the concatenation of the context and query input and thus forms a global attention output. The output of each Transformer encoder is fed in to the next layer, creating an attention hierarchy. The benefit of this construction is that the model has access to the entire query and context at each level allowing both embeddings to learn from each other and removing the long term memory bottleneck faced by RNN based models. BERTQA uses directed coattention between the query and context, as opposed to attending to their concatenation (Figure \ref{fig:newarch}). Our architecture consists of a set of 7 directed coattention blocks that are inserted between the BERT embeddings and the final linear layer before loss calculation. 

The BERT embeddings are masked to produce seperate query and context embedding vectors (Equations \ref{eq:mask1} , \ref{eq:mask2}).
\begin{equation}\label{eq:mask1}
    E_q = E * m_q
\end{equation}
\begin{equation}\label{eq:mask2}
    E_c = E * m_c
\end{equation}
Where E is the contextualized embeddings derived from BERT, m is the mask, and c and q are the context and query respectively.

$E_q$ and $E_c$ are then projected through linear layers to obtain key, value, and query vectors (Equation \ref{eq:proj}).
\begin{equation}\label{eq:proj}
    P_x = W_PE_x^T + b_P \forall P \in {Q,K,V}, x \in {q,c}
\end{equation}
Where Q, K, and V are the query, key and value vectors.

The Q, K, and V vectors are used in scaled dot-product attention (Equation \ref{eq:dpatt}) to create the separate Context-to-Query (C2Q) and Query-to-Context (Q2C) attention vectors. 
\begin{equation}\label{eq:dpatt}
    Coattention(Q,K,V)=softmax(\frac{Q_yK_z^T}{\sqrt{d_k}})V_z
\end{equation}
Where y is q and z  is c for Q2C and y is c and z is q for C2Q.

The C2Q attention vector is summed with the query input and the Q2C attention vector is summed with the context input via a skip connection. Each sum vector is then pushed through a fully connected block and then is added back to the output of the fully connected block via another skip connection. Each sum is followed by a layer-wise normalization. The two resulting 3D C2Q and Q2C vectors are concatenated along the third (embedding) dimension which are combined by two 1D convolutions to create the final 3D vector representing the combination of the C2Q and Q2C attention. We use two convolution layers here so that the concatenated dimension is reduced more gradually so that too much information is not lost. This vector then goes into a final attention head to perform separate self attention pre-processing for the Start logit and End logit prediction layers. The Start logit is generated by a linear layer and the End logit is generated by the output of an LSTM which takes the concatenation of the start span and end span embeddings as an input. We used the BERT architecture code written in Pytorch from the HuggingFace github \cite{github}. We wrote our own code for all of the subsequent architecture.

\subsection{Localized Feature Extraction}
\begin{figure}[h!]
    \centering
    \includegraphics[height=1in]{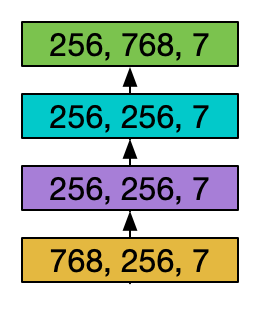}
    \caption{\label{fig:local} Convolutional Layers for Local Attention (in channels, out channels, kernel size)}
\end{figure}

To refine the focus of the attention further, we experimented with convolutional feature extraction to add localized information to the coattention output. We added four convolutional layers within the coattention architecture (Figure \ref{fig:local}). The input to these layers were the BERT embeddings and the outputs were added to the outputs of the multi-head attention layers in the coattention architecture and then layer-wise normalized. This combination of  coattention and local information provides a hierarchical understanding of the question and context. By itself, BERT provides information about the question and context as a unit, while the coattention extracts information from both the question and context relative to each other. The convolutions extract local features within the question and context to add localized information to the attention and embedding meanings. After adding the separate start and end logic, we found that the localized feature extraction did not allow an improvement in the network's learning via an ablation study where we ran the network without the convolutional layers. We speculate that the convolutions prevented improvement beyond a certain F1 score because they are lossy compressors and the information lost by the convolutions might be essential to downstream learning.

\subsection{Skip Connections}
As shown in Figure \ref{fig:newarch}, we have a skip connection from the BERT embedding layer combined with the convolved directed co-attention output (C2Q and Q2C).  We experimented with 3 skip connection configurations: Simple ResNet inspired Skip, Self-Attention Transformer Skip, and a Highway Network. Of these, the Self-Attention Transformer based skip worked best initially. However, when we combined this skip connection with our logit prediction logic, the network was no longer able learn as well. The Simple ResNet inspired skip \cite{resnet} connection solved this issue. It seems that the transformer skip connection followed by the additional transformer encoder blocks that form the beginning of the logit prediction logic processed the BERT embeddings too much and thus lost the benefit of the skip connection. Therefore, we decided to use a Simple ResNet inspired skip alongside the self attention heads for logit prediction. This allows the directed co-attention layers to learn distinct information coming from BERT embeddings via the skip and allows for efficient backpropagation to the BERT layers.

\subsection{Data Augmentation - SQuAD 2.Q}
\begin{figure}[h!]
    \centering
    \includegraphics[height=0.9in]{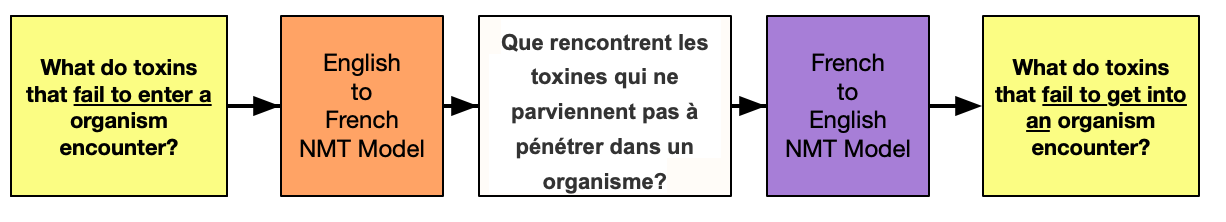}
    \caption{\label{fig:backtrans}Back Translation to augment the SQuAD dataset }
\end{figure}

Inspired by the work presented in \cite{aug} where the authors present a way of generating new questions out of context and after observing the patterns in SQuAD 2.0 we realized there is a lot of syntatic and gramatical variance in the questions written by cloud workers. To help our network generalize better to these variations we decided to augment the dataset by paraphrasing the questions in the SQuAD training set. We applied backtranslation using Google Cloud Translation (NMT) API \cite{google} to translate the sentence from English to French and then do a back translation to English, essentially 2 translations per question (Figure \ref{fig:backtrans}).

 We call our augmented dataset SQUAD 2.Q and make 3 different versions (35\%, 50\%, and 100\% augmentation) alongside code to generate them publicly available on our github \cite{Ank_git}.   

\subsection{Hyperparameter Tuning}
Hyperparameter tuning has been an on-going process for our experiments. Here are the following hyperparameters we have tweaked and tuned for on Bert Base: 
\begin{enumerate}
    \item Number of Directed co-Attention layers - We tried various numbers of layers and we found out that N=7 for the co-attention layers gave us optimal performance while being able to fit the model on 2 GPUs (3 F1 score improvement by itself). 
    \item Max Sequence length - After initial experiments with default sequence length (context + query token) 384, we switched to a sequence length of 512. This gave us a 0.6 F1 improvement on our model. 
    \item Batch Size - Default: 12, We had to use a batch size of 6 for all our experiments due to resource constraints and out of memory issues on the GPU for any larger batch size.  
    \item Number of epochs - Default: 2 On increasing the number of epochs we saw a significant degradation in performance (-3 F1 score), we attribute this to the fact that the model starts to overfit to the training data with high variance and since the batch size is smaller the gradient updates could be noisy not allowing it to optimally converge. 
    \item Learning Rate - Default: 3e-5 We wrote a script to help us find the optimal learning rate using grid search and found the optimal learning rates for SQuAD 2.0 and SQuAD 2.Q respectively for batch size of 6. 
\end{enumerate}
\subsection{BERT Large and Ensembling}
We applied what we learned from the previous five subsections to the large BERT model, which has twice as many layers as the base model. In order to fit this model on our GPU and still use 7 of our coattention layers, we were limited to two examples on the GPU at a time. However, we also found that BERT large requires a larger batch size to get a good performance. As such, we left the batch size 6 as with the base model and used a gradient accumulation of 3 so that only two examples were on the GPU at a time. Additionally, the large model is very sensitive to the learning rate, and the rate of 3e-5 which we used with the smaller model no longer worked. We ran the model on a subset of the data with various learning rates and found that 1.1e-5 to 1.5e-5 works the best for the large model depending on the dataset used (SQuAD 2.0 or SQUAD 2.Q).

After experimenting with multiple combinations of the ideas we described above, we ensembled our three best networks to create our final predictions. The configurations of our three best networks are described in Table \ref{table:config}.

\begin{table*}[!h]
	\centering
    \begin{adjustbox}{width=0.63 \textwidth,center=\textwidth}
	\begin{tabular}{| l | l | l | l | l | l |}
    \hline
    	& BS & GA & LR & Dataset & Arch. Changes   \\
    \hline
    \centering
    Model 1 & 6 & 3 & 1.5e-5 & 2.Q (35\%) & None  \\
    Model 2 & 6 & 3 & 1.2e-5 & 2.Q (35\%) & No LSTM   \\
    Model 3 & 6 & 3 & 1.1e-5 & 2.Q (50\%) & None  \\
    \hline
    \end{tabular}
    \end{adjustbox}
    \caption{Model Configurations; BS = Batch Size, GA = Gradient Accum., LR = Learning Rate}
    \label{table:config}
\end{table*}

We constructed the ensembled predictions by choosing the answer from the network that had the highest probability and choosing no answer if any of the networks predicted no answer.

\section{Results and Analysis} 
Table \ref{table:performance_results} reports the F1 and EM scores obtained for the experiments on the base model. The first column reports the base BERT baseline scores, while the second reports the results for the C2Q/Q2C attention addition. The two skip columns report scores for the skip connection connecting the BERT embedding layer to the coattention output (Simple Skip) and the scores for the same skip connection containing a Transformer block (Transformer Skip). The final column presents the result of the localized feature extraction added inside the C2Q/Q2C architecture (Inside Conv - Figure \ref{fig:local}).

\begin{table*}[!h]
	\centering
    \begin{adjustbox}{width=0.85 \textwidth,center=\textwidth}
	\begin{tabular}{| l | l | l | l | l | l |}
    \hline
    	& Base & C2Q/Q2C & Simple Skip & Transformer Skip & Inside Conv   \\
    \hline
    \centering
    F1                 & 74.15  & 74.34 & 74.81 & 74.95 & \textbf{77.03}   \\
    Has Ans F1 & \textbf{80.62} & 76.30 & 76.13 & 76.35 &         78.47   \\
    No Ans F1          & 68.21  & 72.54 & 73.01 & 73.66 & \textbf{73.83}  \\
    EM                 & 71.09  & 71.56 & 72.11 & 72.07 & \textbf{74.37}   \\
    \hline
    \end{tabular}
    \end{adjustbox}
    \caption{Performance results for experiments relative to BERT base}
    \label{table:performance_results}
\end{table*}

The results presented above verify our hypothesis that adding layers of directed attention to BERT improves its performance. The C2Q/Q2C network produced a significant improvement in the No Answer F1 score while causing a symmetric drop in the Has Answer F1 score. The C2Q/Q2C network attends the context relative to the query and vice versa instead of as a concatenated whole. This method of attention provides more information regarding whether there is an answer to the question in the context than the original BERT attention. The skip connections improved the scores further by adding the BERT embeddings back in to the coattention vectors and providing information that may have been lost by the C2Q/Q2C network in addition to providing a convenient path for backpropagation to the BERT embedding layers. The skip connection containing the transformer provides minimal gains while adding a significant overhead to runtime. Therefore, we built the final convolutional experiments on the Simple Skip architecture. The localized feature extraction within the coattention network produced the best results in the base model, but prevented an improvement in our modified BERT large model.

Table \ref{table:large_results} shows the F1 and EM scores obtained for the experiments on the large model. The models labeled 1, 2, and 3 are described in detail in Section 3.6.
\begin{table*}[!h]
	\centering
    \begin{adjustbox}{width=0.75 \textwidth,center=\textwidth}
	\begin{tabular}{| l | l | l | l | l | l |}
    \hline
    	& BERT Large & Model 1 & Model 2 & Model 3 & Ensemble   \\
    \hline
    \centering
    F1                 & 80.58 & 82.08 & 81.51 & 81.31 & \textbf{83.42}  \\
    Has Ans F1& \textbf{84.39} & 84.36 & 83.53 & 84.38 &         81.09   \\
    No Ans F1          & 77.08 & 79.98 & 79.68 & 78.49 & \textbf{85.56}  \\
    EM                 & 77.74 & 78.81 & 78.24 & 78.00 & \textbf{80.53}  \\
    \hline
    \end{tabular}
    \end{adjustbox}
    \caption{Performance results for experiments relative to BERT large}
    \label{table:large_results}
\end{table*}

Each of the models built on BERT large used our augmented dataset in addition to the coattention architecture, simple skip connection, and separate start and end logit logic. The Model 1 results show that a moderately augmented (35\%) data set helps the training since both unaugmented and highly  augmented (50\%) models did not perform as well. It seems that adding too much augmented data reduces the F1 because the augmented data is noisy relative to the original data. The performance difference between Model 1 and 2 support the use of the LSTM in creating the End logit predictions. The LSTM is successfully combining the information from the Start logit and the End embeddings to provide a good input to the End logit linear layer. The ensemble model performed the best by far due to a significant increase in the no answer F1 which can be attributed to the ensembling method which is biased towards models that predict no answer. 

\begin{figure}[h!]
    \centering
    \begin{subfigure}[h!]{0.5\textwidth}
    \includegraphics[height=3in,trim={0 0 0 1cm},clip]{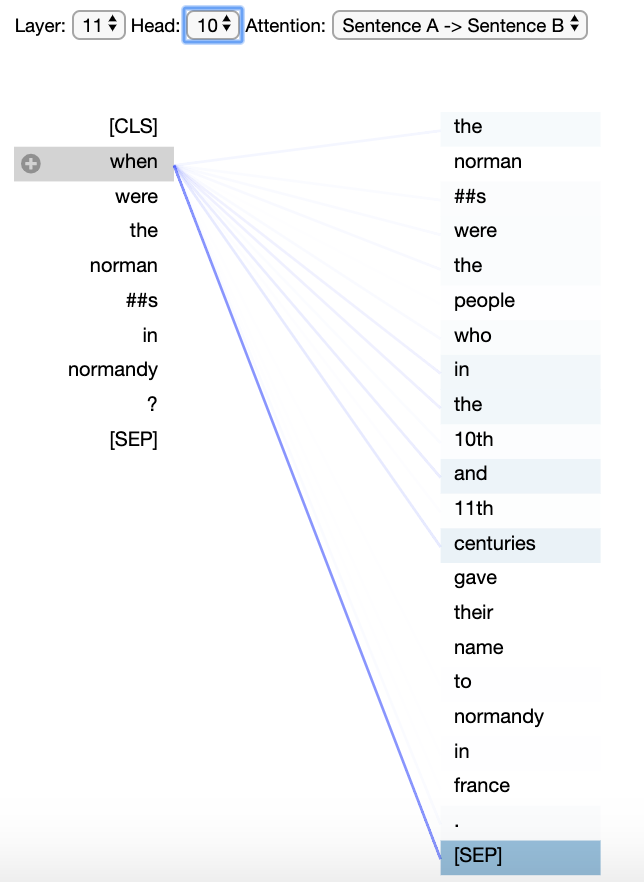}
    \caption{\label{fig:hasans} Question with answer}
    \end{subfigure}
    \begin{subfigure}[h!]{0.4\textwidth}
    \includegraphics[height=3in,trim={0 0 0 1cm},clip]{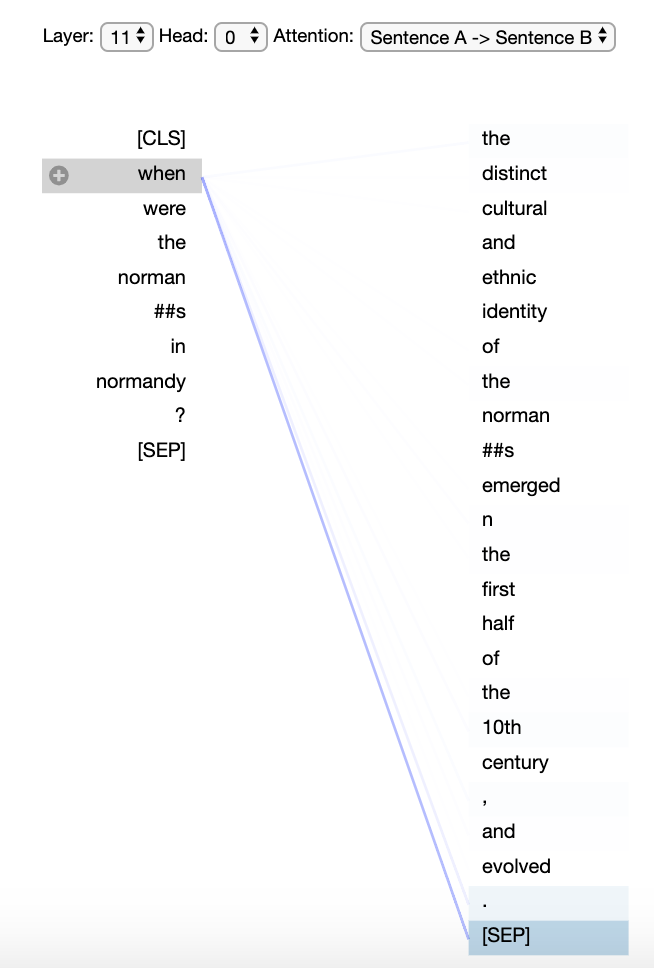}
    \caption{\label{fig:hasans} Question without answer}
    \end{subfigure}
    \caption{\label{fig:vis} Visualization of attention produced by our model}
\end{figure}

We investigated the attention distributions produced by our proposed model by modifying the open source code from BertViz \cite{bertviz} . For the case where the question has an answer in the context (Figure \ref{fig:vis}), the attention heads produce activation around the answer phrase "in the 10th and 11th centuries". In the case where there is no answer in the context, the attention heads produce considerable activation on the [SEP] word-piece which is outside the context span. 

\begin{figure}[h!]
    \centering
    \includegraphics[height=2.5in]{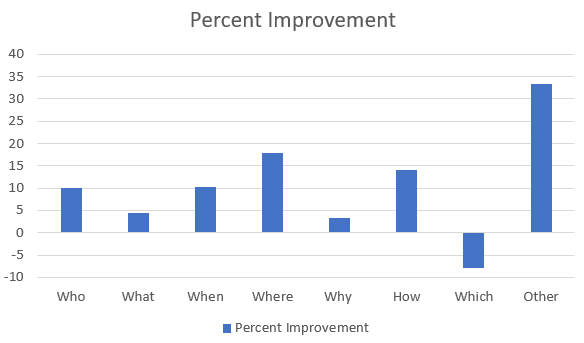}
    \caption{\label{fig:graph}Percent error for different question types }
\end{figure}
As seen in Figure \ref{fig:graph}, we conducted an error analysis over different question types. Note that questions that did not fit into  the 7 bins were classified as "Other". An example of a question in the "Other" category would be an "Is it?" question which is a minority set in SQUAD 2.0. Over the baseline, our model pretty much presents an overall improvement across the board in the different type of questions in the SQuAD 2.0 dev set. In the case of "Which" questions, our model goes wrong 69 times where as the baseline model goes wrong 64 times, a very small numeric difference. However, for the "What" questions the baseline model produces incorrect outputs for 776 examples while our model produces 30 fewer incorrect outputs. The reason for this lapse appears to be related to data augmentation where we observed that many a times "Which" was backtranslated as "What" and vice versa. Thus, the questions in these two classes are mixed and a completely accurate analysis of improvements in these classes is not possible.

\begin{figure}[h!]
    \centering
    \includegraphics[height=1.75in]{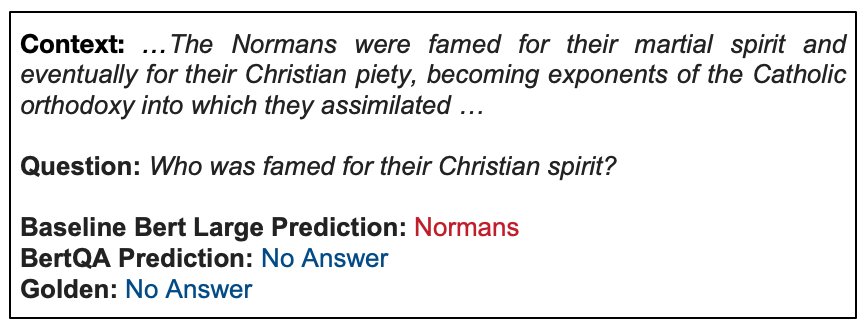}
    \caption{\label{fig:output}Comparison of BERT large and Ensemble performance on an example}
\end{figure}

Figure \ref{fig:output} shows an example cropped context and question that our ensemble model answers correctly while the BERT large model answers incorrectly. It seems that the BERT large model combined the words spirit and Christian to answer this question even thought the word spirit belongs to martial and the word Christian belongs to piety. Our model was able to keep the paired words together and realize that the question has no answer. We believe that our model was able to get the correct answer because of the coattention which is able to keep the words paired together correctly.

Overall, our model has shown marked qualitative and quantitative improvement over the base and large BERT models. Our SQUAD 2.Q dataset helps improve performance by mimicking the natural variance in questions present in the SQUAD 2.0 dataset. BertQA produces a significant improvement in the No Answer F1 by being able to maintain associations between words via coattention, as seen in Figure \ref{fig:output}, and by ensembling our three best models.

\section{Conclusion}
We present a novel architectural scheme to use transformers to help the network learn directed co-attention which has improved performance over BERT baseline. We experimented with several architectural modifications and presented an ablation study. We present SQuAD 2.Q, an augmented dataset, developed using NMT backtranslation which helps our model generalize better over syntatic and grammatical variance of human writing. Our ensemble model gives a ~3.5 point improvement over the Bert Large dev F1. We learned a lot about neural architectural techniques through experimenting with various model configurations. We also learned about how different model components do or don't work together and that some architectural choices like convolutional layers that work so well in computer vision do not necessarily work as well in NLP.

We would like to improve on the quality of data augmentation to limit noise in the dataset and further extend this work to context augmentation as well. Apart from that, we would also like to try recent architectures like Transformer-XL \cite{txl} which has potential to offer additional improvement on HasAns F1 by remembering long term dependencies and evaluate how it scales with our model as a next step. Given sufficient compute resources we would also like to pre-train our C2Q and Q2C layers similar to BERT pre-training to learn deeper language semantics and then fine-tune it on the SQuAD dataset for the task of Question Answering.

We would like to thank the CS224n Team for all the support throughout the course and also thank the folks at Azure for providing us with Cloud credits.

\end{document}